\pdfoutput=1

\documentclass[11pt]{article}

\usepackage{acl}

\usepackage{times}
\usepackage{latexsym}
\usepackage[T1]{fontenc}
\usepackage[a4paper]{geometry}

\usepackage[utf8]{inputenc}

\usepackage{microtype}
\usepackage[pdftex]{graphicx} \pdfcompresslevel=9
\usepackage{xurl}
\usepackage{multirow}

\title{Evons: A Dataset for Fake and Real News Virality Analysis and Prediction}

 \author{Kriste Krstovski,\textsuperscript{1,2}
   Angela Soomin Ryu,\textsuperscript{1} and
  Bruce Kogut\textsuperscript{1,3}\\
  \textsuperscript{1}Columbia Business School, Columbia University \\
  \textsuperscript{2}Data Science Institute, Columbia University\\
   \textsuperscript{3}Department of Sociology, Columbia University\\
  \texttt{\{kriste.krstovski,asr2193,bruce.kogut\}@columbia.edu}}

\begin{document}
\maketitle
\begin{abstract}
We present a novel collection of news articles originating from fake and real news media sources for the analysis and prediction of news virality. Unlike existing fake news datasets which either contain claims or news article headline and body, in this collection each article is supported with a Facebook engagement count which we consider as an indicator of the article virality. In addition we also provide the article description and thumbnail image with which the article was shared on Facebook. These images were automatically annotated with object tags and color attributes. Using cloud based vision analysis tools, thumbnail images were also analyzed for faces and detected faces were annotated with facial attributes. We empirically investigate the use of this collection on an example task of article virality prediction.
\end{abstract}

\section{Introduction}
Fake news articles are widely spread across social media platforms such as Facebook and Twitter. This is mainly due to the fact that social media is gradually becoming the main source of news consumption \cite{shu_ea_2018}. Due to the sharing features that these platforms offer, fake news propagate rapidly and their effects resonate and persist across many users \cite{baly_ea_2018}. The wide spread of fake news in social media has lead to the development of automatic fake news detection approaches \cite{ruchansky_ea_2017, pereze_ea_2018, nguyen_ea_2019, zellers_ea_2019}, to name a few. Developing fake news detection models require annotated collections of fake and real news articles. Most prior work on the creation and annotation of such collections has focused on this task. Significant number of these collections contain claims fact-checked for veracity \cite{vlachos_riedel_2014,wang_2017}. A recent survey of such collections is provided in \citet{guo_ea_2022}. 

On the other hand there exist collections of fake news articles that contain article headline and body text \cite{horne_ea_2018, potthast_ea_2018,xinyi_ea_2020}. Given that these and other existing fake news collections were developed mainly for fake news detection they can't be used for analysing and predicting fake news virality which is the set of tasks of our focus. Recently, \citet{shu_ea_2018} created FakeNewsNet, a collection of $\sim$24k news articles labeled as fake or real using the fact-checking websites PolitiFact \cite{PolitiFact2017} and Gossip Cop \cite{gossipcop}. Articles in this collection are annotated with social engagement information obtained through the Twitter search API. However this collection doesn't include thumbnail images and article descriptions which, along with the headlines, are the only sources of information readers are exposed to on social media platforms regardless of their choices whether to click the link of the shared article or not. 

To address this drawback we present Evons -- a collection of news articles originating from fake and real news media sources where each article has the thumbnail image and description with which it was shared on Facebook. We use the article engagement count on Facebook as an implicit indicator of the article virality. Given that fake news writers profit from advertising revenue rather than subscription fees, the body text of fake news articles (which are only shown after clicking the link) are known to be repetitive and lacking in informational value \cite{horne_adali_2017}. Therefore we believe that these two article components are important for social media sharing. Thumbnail images are annotated with content tags and color attributes while detected faces are annotated with facial attributes. The Evons collection is accessible through \url{https://github.com/krstovski/evons}. We showcase the use of this collection on the task of article virality prediction which we consider as one example task that could be created wit this dataset.

\section{Collection Construction}
\label{sec:data_coll}
The Evons collection contains 92,969 news articles from fake and real news media sources published in the period between January 2016 and December 2017. 
We selected this time period to reflect on the 2016 Presidential election which many believed that fake news had a significant impact on. Across both media sources we focused on news articles originating from the same news sections therefore covering similar or the same set of topics. We also don't consider any article author related information.  
The set of fake news sources was created using information from 3 independent lists of fake news websites that were developed through human curation. It contains only fake news sources that were cross-referenced by at least 2 of the 3 lists. We follow the most widely used definition of fake news as "intentionally and verifiably false, and could mislead readers" by \citet{allcott_gentzkow_2017} and exclude satire and parody websites.

We used the “Questionable Sources” list from Media Bias Fact Check (MBFC) \cite{mbfc} which includes sources with extreme bias, propaganda and conspiracies, and fake news. We filtered the websites to retain only those that are explicitly annotated as “some fake news” or “fake news”, indicating that the source deliberately publishes hoaxes and/or disinformation. 

Our second list is the "Politifact's Fake News Almanac" \cite{PolitiFact2017}. This list was created in partnership with Facebook and includes "fake news" websites which were found to contain \emph{deliberately} false or fake stories that have appeared in people's news feeds on Facebook.

The third list is from the "BS Detector" collection. This is a list of "unreliable or otherwise questionable sources" curated by professionals \cite{Risdal2017}.

After cross-referencing, we obtained 16 fake news sources that appeared in at least 2 of the lists. We then removed sources that were republishing news content from other sources \emph{and} websites that started publishing after the 2016 elections. Our final list contains the following 6 fake news media sources: American Freedom Fighters (AFF), Barracuda Brigade (BB4SP), MadWorldNews (MWN), Puppet String News (PSN), USA Supreme (USAS), and YourNewsWire (YNW).
The set of real news sources was obtained from the readily available "All the news 2.0" dataset \cite{atn} which consists of 18 American mainstream sources. We focused on sources that had “high” or “very high” scores in factual reporting \emph{and} "very slight" or "neutral" political biases according to MBFC. There were 5 such sources in this dataset: The Guardian, National Public Radio (NPR), New York Times (NYT), Reuters, and Washington Post (WP). We use articles published in the same time period as our fake media set. In Table~\ref{tab:media_source_stats} we provide the number of articles across the fake and real news media sources. 

\begin{table}[t]
\begin{center}
\begin{tabular}{ |l|r|r| } 
\hline
\multicolumn{1}{|c|}{Media Source} & \multicolumn{1}{|c|}{\# of Articles}\\ 
\hline
AFF & 7,536  \\
BB4SP & 2,792 \\
MWN & 11,315 \\
PSN & 6,576 \\
USAS & 3,038\\
YNW & 11,519 \\
\hline

Total from fake & 42,776\\
\hline
\hline
The Guardian & 9,811 \\
NPR & 11,813 \\
NYT & 5,439  \\
Reuters & 14,993 \\
WP & 8,137\\
\hline
Total from real & 50,193\\
 \hline
 \hline
 Total & 92,969\\
 \hline 
\end{tabular}
\end{center}
\caption{Number of articles in the Evons collection.}
\label{tab:media_source_stats}
\end{table}

We used the webpreview\footnote{https://pypi.org/project/webpreview} package for extracting thumbnail images. These images come from the thumbnails that are carefully curated by the news producers. They decide what title, description, and thumbnail image would be the most effective in achieving their goal, whether it is to best represent the content or simply attract the most engagement for larger advertising revenue. With this package we also extract article description which is the text that appears as preview when the article is shared. 

All articles contain a thumbnail image except for USAS and BB4SP were 0.1\% and 11.1\% of the articles don't have thumbnails. Thumbnail images are either a picture or a logo of the news media source. Table~\ref{tab:thumbnail_stats} gives statistics of the number of real and fake articles with and without thumbnail images. Unlike real news articles where a small percentage of them had the media source logo as the thumbnail image, fake news articles always used pictures as thumbnails. 

\begin{table}[ht]
\begin{center}
\begin{tabular}{ |l|r|r|r|}
 \hline
\multicolumn{1}{|c|}{Thumbnail Type} & \multicolumn{1}{|c|}{Real}  & \multicolumn{1}{|c|}{Fake}    & \multicolumn{1}{|c|}{Total} \\
\hline
Picture &    48,592  & 42,464 & 91,056 \\
Logo    &         1,601  &  0 & 1,601 \\
None    &       0  &    312 & 312 \\
\hline
\end{tabular}
\end{center}
\caption{Thumbnail statistics.}
\label{tab:thumbnail_stats}
\end{table}

\subsection{Engagement Count}
\label{sec:engagement_count}
A commonly used measure for virality by marketing and communication researchers is how many times a piece of information is shared \cite{berger_milkman_2012, scholz_ea_2017}. Here we use Facebook engagements as a proxy of how much attention the post generated. Facebook engagements is the sum of the number of Facebook shares, likes, and comments. Facebook provides the numbers received by an URL through the Facebook sharing debugger (FSD) \cite{fsd}. 
Since FSD works on individual URLs we used the Shared Count API \cite{sharedcount} to automate the process of fetching these numbers for multiple articles, except for articles from USAS which was blacklisted on Facebook. For this website we used BuzzSumo \cite{BuzzSumo} which is another third-party measurement dashboard that fetches data from FSD. Both platforms do not provide nor maintain any user related information and have been used in the past across an array of research topics \cite{xu_hao_2018, xu_2019, obiala_ea_2021, rhodes_2022}.
In Table~\ref{tab:engagement_stats} we provide engagement statistics.

\begin{table}[t]
\begin{center}
\begin{tabular}{ |l|r|r| } 
 \hline
\multicolumn{1}{|c|}{Engagement Statistics} & \multicolumn{1}{|c|}{Real}  & \multicolumn{1}{|c|}{Fake}   \\ 
\hline
Min \# of engagements  &         0   &       0 \\
Max \# of engagements  &     4.78m   &  1.08m  \\
Mean \# of engagements &     6.73   &   1.58 \\
\hline
\end{tabular}
\end{center}
\caption{Engagement statistics.}
\label{tab:engagement_stats}
\end{table}

\subsection{Image Annotation}
We performed two types of automatic image annotation. Using Microsoft Azure \cite{azure_computer_vision} images are analyzed for visual features and color schemes. With the Amazon Rekognition platform \cite{rekognition_facial_analysis} images are analyzed for the presence of faces and detected faces were annotated with facial attributes. Accuracy of both platform on these annotation tasks have been extensively evaluated and confirmed in the past across a variety of image types which include images commonly used as thumbnails \cite{kyriakou2019fairness, liu2020image, malone_burns_2021}. 
\subsubsection{Object Detection and Tagging}
Images are automatically annotated with content tags such as objects, living beings, scenery, and actions. There were 5,160 distinct tags identified. Articles originating from fake media sources had 3,670 distinct tags with 379 being unique to fake. Real sources contained 4,781 distinct tags with 1,490 unique to real. Table~\ref{tab:image_stats} shows image tag statistics. Table~\ref{tab:object_tags} shows the top 10 most frequent tags discovered across all media sources, unique to fake, and real news sources. 

\begin{table}[t]
\begin{center}
\begin{tabular}{ |l|r|r| } 
\hline
\multicolumn{1}{|c|}{Image Tag Statistics} & \multicolumn{1}{|c|}{Real}  & \multicolumn{1}{|c|}{Fake} \\
\hline
Min \# of tags    &          0   &   0    \\
Max \# of tags    &        99   &     86  \\
Mean \# of tags  &     9.47   &  9.08  \\
\hline
\end{tabular}
\end{center}
\caption{Image tag statistics. }
\label{tab:image_stats}
\end{table}

\begin{table*}[ht]
\begin{center}
\begin{tabular}{ |l|l|l| } 
 \hline
\multicolumn{1}{|c|}{All} &  \multicolumn{1}{c|}{Unique to Real} & \multicolumn{1}{c|}{Unique to Fake}\\
\hline
 1. person       &  1. salad & 1. photo caption        \\
 2. clothing     &  2. minimalist & 2. television presenter  \\
 3. human face   &  3. raquet sport & 3. thong                  \\ 
 4. man          &  4. racketlon & 4. shout                \\
 5. text         &  5. piece de resistance & 5. g-string              \\
 6. outdoor      &  6. tennis player & 6. f-15 eagle             \\
 7. suit         &  7. soft tennis & 7. salumi              \\    
 8. indoor       &  8. modern & 8. salami                 \\
 9. smile        &  9. professional boxing & 9. ciauscolo             \\
10. tie          & 10. camera lens & 10. ostrich                \\   
\hline
\end{tabular}
\end{center}
\caption{Top 10 most frequent tags across all media sources, unique to real, and fake news sources.}
\label{tab:object_tags}
\end{table*}

\subsubsection{Color Schemes}
Thumbnail images are automatically annotated with three color attributes: dominant foreground and background color, and a set of dominant colors across the whole image. There are 12 colors used: black, blue, brown, gray, green, orange, pink, purple, red, teal, white, and yellow. Dominant background and foreground colors can take on a single value. Thumbnails are also annotated with accent color, which is the most vibrant color in the image, and whether the image is in black and white (bw). In Appendix~\ref{appendix:dominantcolors} we provide  summary of the colors present as dominant attribute in thumbnail images.

\subsubsection{Facial Analysis}
Detected faces are annotated with a bounding box and the following attributes: person's gender, whether the person is smiling, wearing eyeglasses or sunglasses, has a mustache or eyes open, brightness, and sharpness. We also obtain the emotions that appear to be expressed on the face which include: fear, sad, happy, calm, angry, confused, surprised, and disgusted. Table~\ref{tab:face_stats} provides face statistics. In Appendix~\ref{appendix:dominantemotion} we show the distribution of dominant face emotions. 

\begin{table}[t]
\begin{center}
\begin{tabular}{ |l|r|r| } 
\hline
\multicolumn{1}{|c|}{Face Statistics} & \multicolumn{1}{|c|}{Real}  & \multicolumn{1}{|c|}{Fake} \\
\hline
\% of images with face/s &    74.26 & 77.08 \\ 
Mean \# of faces per image &    3.31 & 2.74  \\
\hline
\end{tabular}
\end{center}
\caption{Face statistics. }
\label{tab:face_stats}
\end{table}

\section{Example Task}
 We use the task of predicting article virality as an example task (out of many different tasks) that could be constructed using the Evons collection. The example task is a multi-class classification problem which we created by dividing articles from fake and real news media sources into two groups based on their engagement count: real-low, real-high, fake-low, and fake-high. We use the median number of engagements to create almost equal groups of real and fake articles with low and high number of engagements. We empirically investigate how well do various approaches, which we consider as baselines, perform on this task.
 
\subsection{Experimental Setup and Results}
\label{experimental_setup}
The task dataset consists of articles with pictures as thumbnails where the picture contained at least one tag and face. There are 68,793 such articles out of which 36,072 come from real and 32,721 from fake media sources. Articles are represented using two sets of textual features and three sets of image features, one for each of the three image annotation types. For the textual features we use tf-idf values computed over the words of article titles and descriptions. The title feature vector contains 29,745 words and the description feature vector with 43,861 words. Combining both we obtain a vocabulary of 49,792 words. Thumbnail images were represented with 3,526 features: 3,471 object tags, 42 color and 13 facial. Color features include accent color, dominant color attributes, and bw indicator. Facial features include number of faces, person smiling, gender, brightness, sharpness, and facial emotions. Facial features were weighted based on the size of the bounding box area of the detected face. In Appendix~\ref{appendix:facial_features} we provide details on the weighing approach used. For features that are indicator variables we use the confidence score as a feature value. 

We evaluated 6 different classification models: logistic regression (LR), SVM, multilayer perceptron (MLP), Bidirectional LSTM \cite{lstm} (Bi-LSTM), XLNet \cite{yang_ea_2019}, and RoBERTa \cite{liu_ea_2019}; using a 90/10 split of our dataset. We used the scikit-learn \cite{pedregosa_ea_2011} implementation of LR and SVM. MLP consists of three fully-connected layers containing 256 and 8 nodes in the first two layers with ReLU. The last layer is a 4 nodes with SoftMax activation. Bi-LSTM consists of a 64 dimensional embedding representation layer, a fully connected layer with ReLU, and an output layer as in MLP. Both NNs were implemented in Keras \cite{gulli_ea_2017}. We used the simpletransfomers \cite{simpletransformers} implementation of XLNet and RoBERTa with maximum sequence length of 256.  Table~\ref{tab:share_classify} shows performance comparison results across all models using different feature representations and combinations of them. For ease of interpretability we use accuracy. Thumbnail images were represented using all image generated features. RoBERTa with all feature types performs best. 
While across most models incorporating image features helps we don't observe substantial accuracy improvement over textual features. We believe that this could be significantly improved with image feature analysis and exploring feature selection approaches.  

\label{sect:result_analysis}
\begin{table*}[ht]
\begin{center}
\begin{tabular}{|l|r|r|r|r|r|r|} 
\hline
 \multicolumn{1}{|c|}{\multirow{2}{*}{Feature}} & \multicolumn{6}{c|}{Accuracy} \\
\cline{2-7} 
    & LR     & SVM    & MLP  & Bi-LSTM & XLNet & RoBERTa   \\
\cline{2-7}
\hline
Title (T)        & 0.632  & 0.608  & 0.643 & 0.632 & 0.731 & 0.751 \\
Description (D)        & 0.674  & 0.631  & 0.680 & 0.687 & 0.760 & 0.773  \\
T+D             & 0.694  & 0.655  & 0.718 & 0.691 & 0.801 & 0.807  \\
\hline
T+D+Tag         & 0.701  & 0.661  & 0.719 & 0.712 & 0.793 & 0.808  \\
T+D+Color       & 0.701  & 0.658  & 0.716 & 0.688 & 0.781 & 0.801  \\
T+D+Facial      & 0.697  & 0.655  & 0.716 & 0.688 & 0.794 & 0.802  \\
\hline
All             & 0.703  &  0.666 & 0.714 & 0.683 & 0.791 & 0.810  \\
\hline
\end{tabular}
\end{center}
\caption{Accuracy results across various baseline models on the example task of article virality prediction.} 
\label{tab:share_classify}
\end{table*}

\section{Conclusion}
 \label{sect:conlusion}
We presented Evons - a collection of news articles originating from fake and real media sources where articles are annotated with a Facebook engagement count, thumbnail image and article description. Thumbnails are automatically annotated with object tags, color and facial attributes. We demoed the collection use on an article virality prediction task and established baselines using 6 models. In the future we plan to use Evons to explore various approaches for selection of image features and combination with text that would further help improve accuracy on this task. 
\section{Ethics}
\label{sect:ethics}
Creating the Evons collection involved collecting news articles from various online media sources, extracting thumbnail images using the webpreview package, and obtaining Facebook engagement counts through the SharedCount API and the BuzzSumo platforms. Throughout the creation process we made sure that no author metadata or user identifying information was collected. Therefore our collection does not contain any information that names or uniquely identifies individual people. Both Facebook engagement counts platforms do not provide any user related information. While news articles across various online media sources do provide article author information in our collection process we ignored this information.   

We don't foresee any potential risks that may arise from the creation of our collection especially in terms of identifying potential stakeholders that may benefit from this collection while harming others. To the best of our knowledge all of our collected data is in the public domain and is not copyrighted. 

For our thumbnail image annotations we relied on two image annotation platforms: Microsoft Azure and Amazon Rekognition. One limitation of our work may arise from the fact that we don't know whether the models that are part of these platforms contain any type of bias and if so to which extent bias is present.

\bibliographystyle{acl_natbib}
\bibliography{coling}

\clearpage

\appendix

\section{Dominant Colors}
\label{appendix:dominantcolors}
\begin{figure}[h]
   \centering
   \includegraphics[width=\columnwidth]{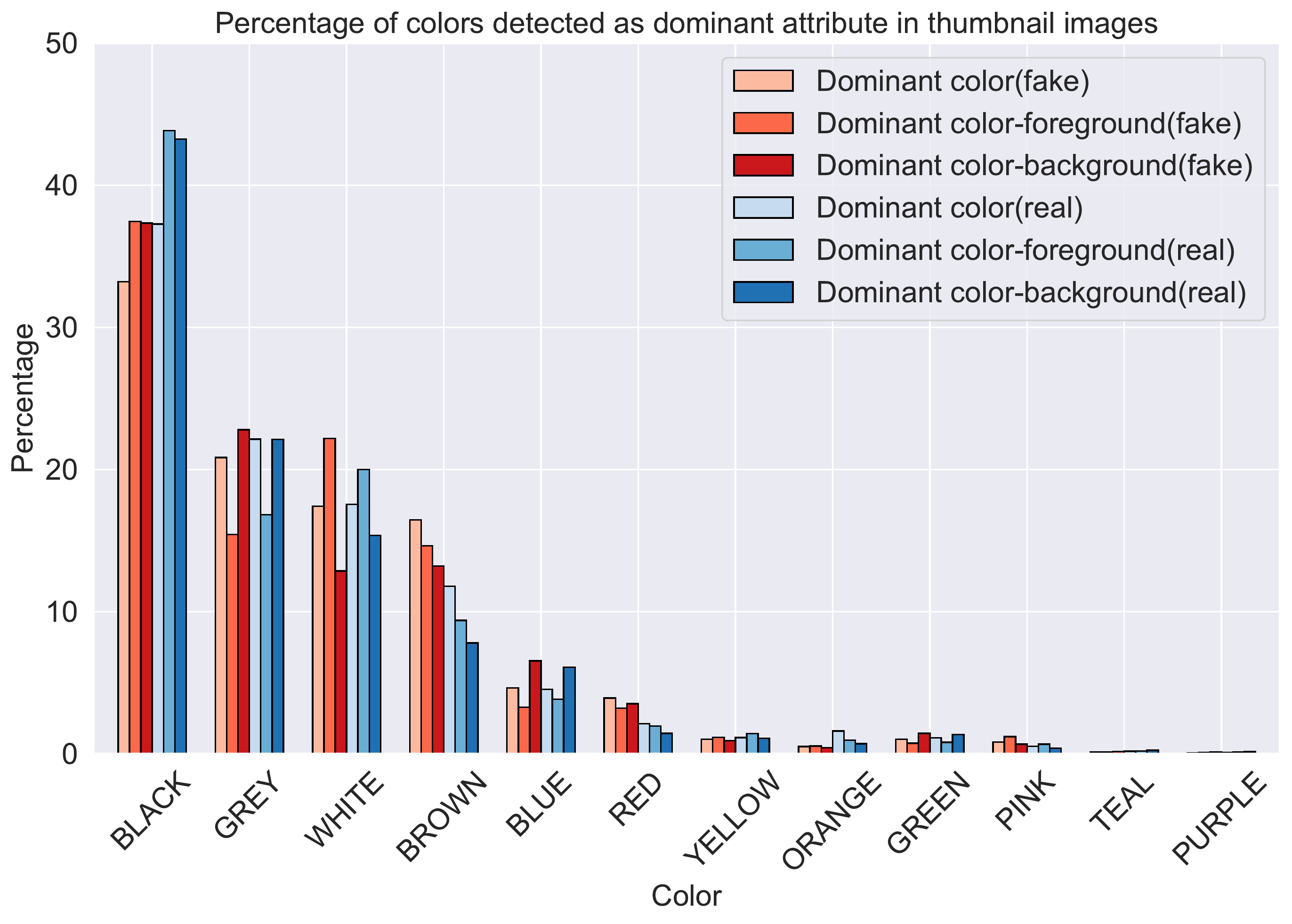}
   \caption{Percentage of color present as dominant attribute in thumbnail images.}
   \label{fig:colorschemes}
\end{figure}

Shown in Figure~\ref{fig:colorschemes} are bar plots of the percentage of colors present as dominant attribute in thumbnail images.

\section{Dominant Emotions}
\label{appendix:dominantemotion}
\begin{figure}[h]
   \centering
   \includegraphics[width=\columnwidth]{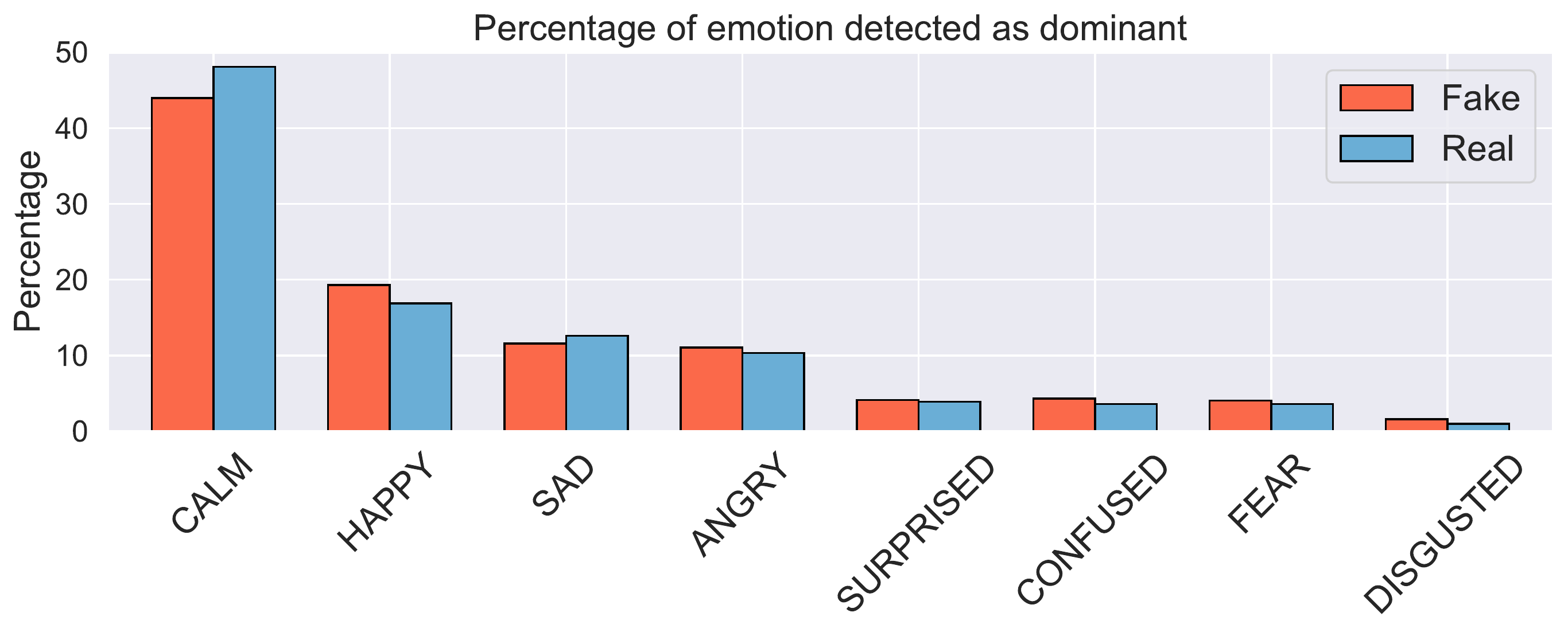}
   \caption{\% of emotion detected as dominant in faces.}
   \label{fig:dominantemotion}
\end{figure}

Shown in Figure~\ref{fig:dominantemotion} are bar plots of the percentage of emotion detected as dominant on faces found in thumbnail images.

\section{Facial Features}
\label{appendix:facial_features}
Facial features across thumbnail images where weighted based on the bounding box area of the detected face. The bounding box area is the product of the bounding box width and height. Given a bounding box area $B_{ij}$ of the $j$th face in image $i$ and a set of $k$ features $F_{jk}$ detected on that face, the weighted facial features for image $i$, $W_{ik}$ are computed as:
\begin{eqnarray}\label{eq:weighted}
	W_{ik}=\sum_{j=1}^J{B_{i,j}F_{j,k}}
\end{eqnarray}	
\end{document}